\documentclass[conference]{IEEEtran}
\usepackage{cite}
\usepackage{amsmath,amssymb,amsfonts, bm}
\usepackage{algorithmic}
\usepackage{graphicx}
\usepackage{textcomp}
\usepackage{xcolor}
\usepackage{url}
\usepackage{float}

\usepackage{doi}

\begin{document}
%
%
\title{Learning Characteristics of Reverse Quaternion Neural Network 
\thanks{Part of this work has been done while the first author, Yamauchi, was a graduate student at Rikkyo University. }
}


%

%
\author
{\IEEEauthorblockN{Shohgo Yamauchi}
\IEEEauthorblockA{\textit{The Asashi Shimbun Company} \\
Tokyo, Japan \\
yamauchi-s1@asahi.com}
\and
\IEEEauthorblockN{Tohru Nitta}
\IEEEauthorblockA{\textit{Tokyo Woman's Christian University} \\
Tokyo, Japan \\
tnitta@lab.twcu.ac.jp}
\and
\IEEEauthorblockN{Takaaki Ohnishi }
\IEEEauthorblockA{\textit{Rikkyo University} \\
Tokyo, Japan \\
ohnishi@rikkyo.ac.jp}
}

\maketitle              

\begin{abstract}
The purpose of this paper is to propose a new neuron structure for multilayer feedforward quaternion neural networks called Reverse Quaternion Neural Network, which utilizes the non-commutative property of quaternion multiplication, and to clarify its learning characteristics.
While quaternion neural networks have been used in various fields, there have been no research reports on the characteristics of multi-layer feedforward quaternion neural networks that apply weights in the reverse direction.
In this paper, we examined the learning characteristics of the Reverse Quaternion Neural Network through two experimental studies: line rotation tasks and three-dimensional object rotation tasks using ModelNet10. Through comparative analysis with conventional models, we investigated both the generalization capabilities and internal learning dynamics.

Our results demonstrated that while the Reverse Quaternion Neural Network achieved performance comparable to existing models, it exhibited distinct characteristics in both rotational generalization capability and learning behavior, as evidenced by its gradient dynamics and feature acquisition process.
\end{abstract}

%
%
%
\section{Introduction}
%
In recent years, machine learning technology has made tremendous progress with the rise of deep learning. In particular, it has achieved remarkable success in fields such as image recognition, natural language processing, and speech recognition, and is still being actively researched.
A high-dimensional neural network \cite{nit2009,mandic2009,hirose2013,parcollet2020,lee2022} is a type of neural network that uses numbers of two or more dimensions, such as complex numbers and quaternions, to represent the parameters of the neural network . 
High-dimensional neural networks can deal with hypercomplex-valued signals naturally. 
It is well-known that complex-valued neural networks and quaternion neural networks require fewer parameters (weights, biases) and have several times faster learning speeds than usual real-valued neural networks 
\cite{nit1991,nit1993,complexNN,QuaternionBP}. 

%
Recent studies have actively explored the application of high-dimensional neural networks in areas such as speech recognition and image processing. 
Zhu et al.\cite{zhu2019quaternion} have extended the convolutional neural networks (CNN) \cite{lecun1998gradient} to quaternions. This approach represents the relationship between RGB colors in an image through the rotation of quaternions in the imaginary parts ({\sf i}, {\sf j}, and {\sf k} axes), and has shown higher accuracy in color image processing compared to real-valued convolutional neural networks. 
The quaternion convolutional neural network, proposed by Parcollet et al. \cite{parcollet2018quaternion}, introduces a model that divides the feature map into individual components of quaternions and convolves them. This approach has demonstrated superior recognition capabilities in speech recognition tasks compared to real-valued CNNs.

%
A quaternion is a mathematical concept introduced by Hamilton \cite{Quaternion} in 1843. 
A quaternion consists of a real part and three imaginary parts ({\sf i}, {\sf j}, {\sf k}), each representing an independent dimension. This characteristic makes quaternions particularly suitable for representing rotations in a three-dimensional space. 
Furthermore, quaternions possess the property that the commutative law of multiplication does not apply, meaning that the order of multiplication significantly affects the outcome. 
As described below, neural networks utilizing quaternions exhibit different characteristics compared to real-valued and complex-valued neural networks in this meaning. 

%
A quaternion neural network is a neural network model where all parameters are represented as quaternions, enabling the network using quaternions to represent data in higher dimensions, and it particularly excels in representing rotations and orientations in three-dimensional space. 
By utilizing the rich expressiveness of quaternions, quaternion neural networks can capture spatial data features that traditional neural networks may miss, especially in fields where spatial information is crucial, such as 3D graphics and robotics.

%
In the context of 3D spatial transformations, Matsui et al. \cite{matsui2004quaternion} conducted an experiment comparing three tasks - scaling, parallel translation, and rotation - using geometric object data with the Quaternion Multi-layer Perceptron (QMLP) and the real-valued MLP (Multi-layer Perceptron). 
The results showed that while the real-valued MLP failed to learn the transformations, the QMLP successfully learned them and performed perfectly in all three tasks.

%
The existence of two types of quaternion neurons, based on non-commutativity, was pointed out in \cite{QuaternionBP}. 
Yoshida et al. \cite{Hopfield-type_QNN} investigated the existence conditions of the energy functions for the Hopfield-type recurrent quaternion neural network and the one with weights applied in reverse order . 
As a result, they clarified that there is no difference between the usual Hopfield-type recurrent quaternion neural network model and the quaternion neural network with weights applied in reverse order.

%
In this paper, we proposed a new neuron structure for multi-layer feedforward quaternion neural networks called the Reverse Quaternion Neural Network (RQNN), which utilizes the non-commutative property of quaternions. The RQNN enables the construction of neural networks with different characteristics by altering the order of quaternion multiplication. This approach captures information and provides expressive capabilities that conventional quaternion neural networks cannot achieve.
Specifically, we conducted experiments based on line rotation and three-dimensional object rotation tasks to evaluate the generalization capability and learning characteristics of the RQNN compared to conventional multilayer feedforward quaternion neural networks. Our experimental results revealed that the RQNN not only acquires rotational representations distinct from those of conventional quaternion neural networks but also exhibits unique learning characteristics.
%
\section{Reverse Quaternion Neural Network Architecture}
In this section, we formulate the RQNN architecture, 
and derive the learning algorithm. 

%
\subsection{Quaternion and Its Properties}
First, we describe the quaternion and its properties. 

The quaternion $q \in \mathbb{H}$ is defined as
\begin{align}
    q = a + b{\sf i} + c{\sf j} + d{\sf k} 
    \in \mathbb{H} \quad \left(a, b, c, d \in \mathbb{R} \right), 
\end{align}
where $a$ is the real part, $b, c, d$ are the imaginary parts, 
$\mathbb{H}$ is the set of quaternions, and 
$\mathbb{R}$ is the set of real numbers. 
${\sf i}, {\sf j}, {\sf k}$ are independent imaginary units, and the following arithmetic rules apply. 
\begin{eqnarray}
    & & {\sf i}^2 = {\sf j}^2 = {\sf k}^2 = {\sf i}{\sf j}{\sf k} = -1. \\
%
    & & {\sf i}{\sf j} = -{\sf j}{\sf i} = {\sf k},\quad 
    {\sf k}{\sf i} = -{\sf i}{\sf k} = {\sf j}, \quad 
    {\sf j}{\sf k} = -{\sf k}{\sf j} = {\sf i}.
\end{eqnarray}
For the two quaternions $q_1 = a_1 + b_1{\sf i} + c_1{\sf j} + d_1{\sf k} 
\in \mathbb{H}$ and $q_2 = a_2 + b_2{\sf i} + c_2{\sf j} + d_2{\sf k} \in \mathbb{H}$, 
their multiplication is calculated as follows: 
\begin{align} \label{multple}
    q_1q_2 &= (a_1a_2 - b_1b_2 - c_1c_2 - d_1d_2) \notag \\
    &\quad + (a_1b_2 + b_1a_2 + c_1d_2 - d_1c_2) \ {\sf i} \notag \\
    &\quad + (a_1c_2 - b_1d_2 + c_1a_2 + d_1b_2) \ {\sf j} \notag \\
    &\quad + (a_1d_2 + b_1c_2 - c_1b_2 + d_1a_2) \ {\sf k}. 
\end{align}
The reverse multiplication $q_2q_1$ is calculated as 
\begin{align} \label{re_multple}
        q_2q_1 &= (a_2a_1 - b_2b_1 - c_2c_1 - d_2d_1) \notag \\
        &\quad + (a_2b_1 + b_2a_1 + c_2d_1 - d_2c_1) \ {\sf i} \notag \\
        &\quad + (a_2c_1 - b_2d_1 + c_2a_1 + d_2b_1) \ {\sf j} \notag \\
        &\quad + (a_2d_1 + b_2c_1 - c_2b_1 + d_2a_1) \ {\sf k}.
\end{align}
From Eqs. (\ref{multple}) and (\ref{re_multple}), it is evident that the multiplication of quaternions is \textit{non-commutative}. 
The conjugate quaternion of $q = a + b{\sf i} + c{\sf j} + d{\sf k}$, denoted as $\overline{q}$, is defined as 
\begin{align}
    \overline{q} = a -b {\sf i} -c {\sf j} -d {\sf k}. 
\end{align}

%

\subsection{A Reverse Quaternion Neuron}
\begin{figure}[H]
    \centering
    \includegraphics[width=100mm, height=50mm]{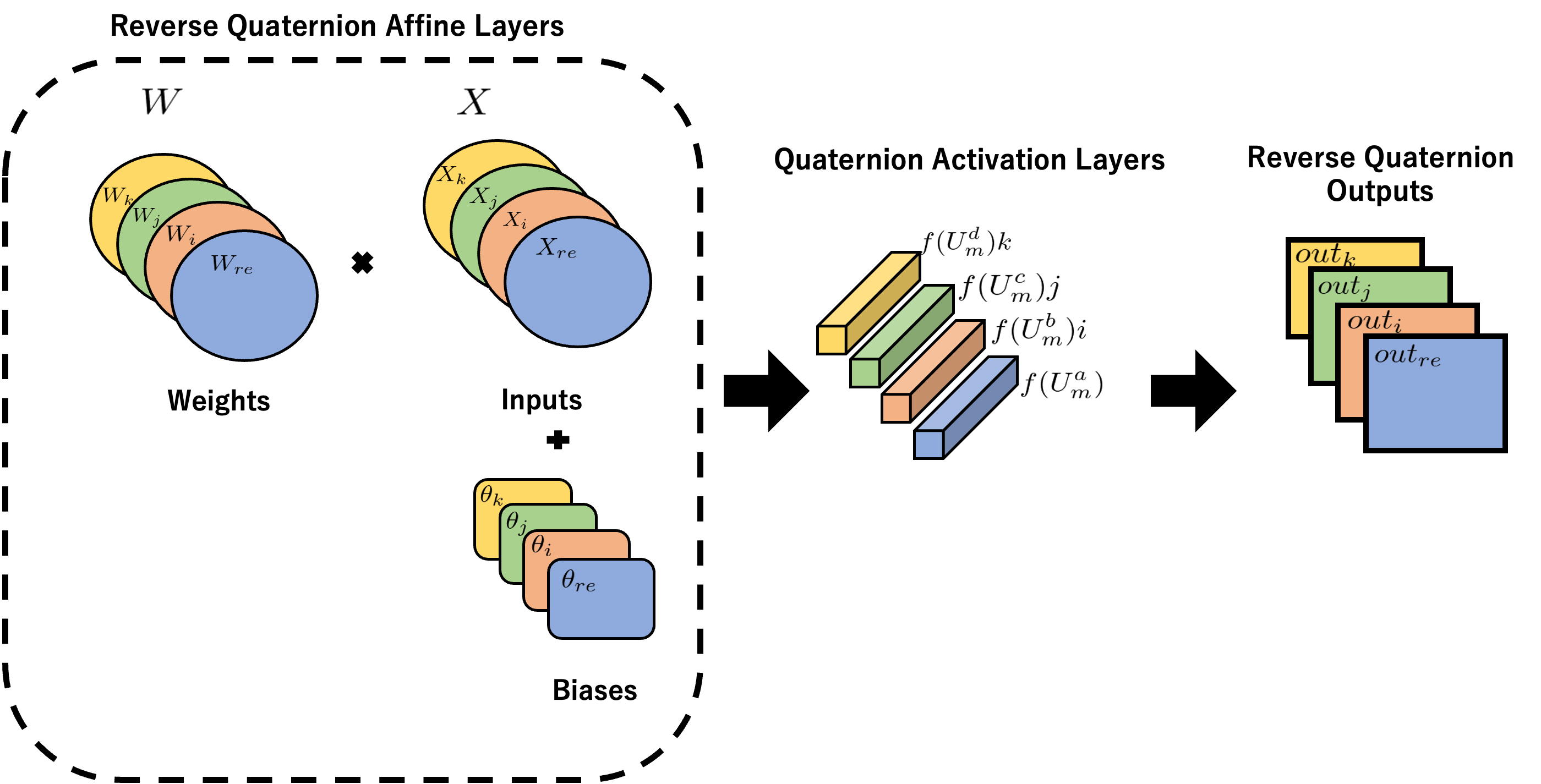}
    \caption{In the Reverse Quaternion Neural Network, unlike the usual quaternion neural network, weights are applied in reverse order. }
    \label{affine}
\end{figure}

A neuron used in the RQNN is as follows. We call it a \textit{reverse quaternion neuron}. 
The input signals, weights, biases, and output signals are all quaternions.
The net input $U_n$ to neuron $n$ is defined to be  
\begin{equation} 
\label{eqn1}
U_n = \sum_m W_{nm} S_m  + T_n \in \mathbb{H},  
\end{equation}
where 
$S_m \in \mathbb{H}$ is the quaternion input signal coming from the output of neuron $m$, 
$W_{nm} \in \mathbb{H}$ is the quaternion weight connecting neurons $m$ and $n$, 
and $T_n \in \mathbb{H}$ is the quaternion bias of neuron $n$. 
Note here that the order of multiplication of a weight and an input signal is 
\textit{WEIGHT times INPUT SIGNAL} (Eq. (\ref{eqn1}), Fig. \ref{affine}). 
%
In a usual quaternion neuron, the order of multiplication of a weight and an input signal is 
\textit{INPUT SIGNAL times WEIGHT} \cite{QuaternionBP,parcollet2020}. 
This is why it is called a reverse quaternion neuron. 
To obtain the quaternion output signal, 
convert the net input $U_n$ into its four parts as follows. 
\begin{equation}
\label{eqn2}
U_n = u_1 + u_2{\sf i} + u_3{\sf j} + u_4{\sf k} \in \mathbb{H}. 
\end{equation}
Although the activation function of a quaternion neuron is a quaternion function generally, 
we use the following activation function, known as \textit{split-type}: 
\begin{eqnarray}
\label{eqn3}
f(x) &=& f_R(x_1) + f_R(x_2){\sf i} + f_R(x_3){\sf j} + f_R(x_4){\sf k} 
\end{eqnarray}
where $x = x_1 + x_2{\sf i} + x_3{\sf j} + x_4{\sf k} \in \mathbb{H}$ and 
$f_R(x): \mathbb{R} \rightarrow \mathbb{R}$ is a real-valued activation 
function applied to each component. 
Therefore, the output value of the neuron $n$ is given by $f(U_n)$.

%
\vspace{0.5cm}


%
\subsection{Reverse Quaternion Neural Network}
\label{RQNN}
The multi-layer feedforward reverse quaternion neural network consists of 
the reverse quaternion neurons described in the previous section. 

%
For the sake of simplicity, consider a three-layer feedforward reverse quaternion 
neural network. 
%
We use $w_{ml} = w^a_{ml}+w^b_{ml}{\sf i} + w^c_{ml}{\sf j} + w^d_{ml}{\sf k} 
\in \mathbb{H}$ \ 
for the weight between the input neuron $l$ and the hidden neuron $m$,  
$v_{nm}=v^a_{nm}+v^b_{nm}{\sf i} + v^c_{nm}{\sf j} + v^d_{nm}{\sf k} \in \mathbb{H}$ \ 
for the weight between the hidden neuron $m$ and the output neuron $n$, \ 
$\theta_m=\theta^a_m+\theta^b_m{\sf i} + \theta^c_m{\sf j} + \theta^d_m{\sf k} 
\in \mathbb{H}$ \ 
for the bias of the hidden neuron $m$, \ 
$\gamma_n = \gamma^a_n + \gamma^b_n{\sf i} + \gamma^c_n{\sf j} + \gamma^d_n{\sf k} \in \mathbb{H}$ \ 
for the bias of the output neuron $n$. \ 
Let $I_l = I^a_l + I^b_l{\sf i} + I^c_l{\sf j} + I^d_l{\sf k} \in \mathbb{H}$ denote 
the input signal to the input neuron $l$, \ 
and let $H_m = H^a_m + H^b_m{\sf i} + H^c_m{\sf j} + H^d_m{\sf k} \in \mathbb{H}$ 
and $O_n = O^a_n + O^b_n{\sf i} + O^c_n{\sf j} + O^d_n{\sf k} \in \mathbb{H}$ \ 
denote the output signals of the hidden neuron $m$, \ 
and the output neuron $n$, \ respectively. \  
%
%
%
Let 
$\delta_n = \delta^a_n + \delta^b_n{\sf i} + \delta^c_n{\sf j} + \delta^d_n{\sf k}  
= T_n - O_n \in \mathbb{H}$ denote the error between $O_n$ \ and the 
target output signal $T_n = T^a_n + T^b_n{\sf i} + T^c_n{\sf j} + T^d_n{\sf k} 
\in \mathbb{H}$ \ 
of the pattern to be learned for the output neuron $n$. \ 
We define the square error for the pattern $p$ as
$E_p = (1/2) \sum_{n=1}^N \vert \delta_n \vert^2$, 
where $N$ is the number of output neurons, and 
$\vert x \vert := \sqrt{x_1^2 + x_2^2 + x_3^2 + x_4^2}$ for 
$x=x_1 + x_2{\sf i} + x_3{\sf j} + x_4{\sf k} \in \mathbb{H}$. 

We used here the same notations as those in \cite{QuaternionBP} for inputs, weights, biases, 
outputs, errors and target output signals.

\subsection{Learning Algorithm}
Next, we derive a learning rule for the RQNN described in Section \ref{RQNN}. 

%
For a sufficiently small learning constant $\varepsilon > 0$, and using a 
steepest descent method, 
we can show that the weights and the biases 
should be modified according to the following equations  as in \cite{QuaternionBP}. 
In this case, the amount of the correction for the RQNN is calculated 
by obtaining the partial derivatives of the real and imaginary parts separately. 
This is a variant of the well-known backpropagation learning algorithm 
using quaternions \cite{rumel}. 
{\small
\begin{eqnarray}
\label{eqn2-1}
   \Delta v_{nm}  &:=& 
   \Delta v_{nm}^a
 + \Delta v_{nm}^b{\sf i} 
 + \Delta v_{nm}^c{\sf j} 
 + \Delta v_{nm}^d{\sf k} \nonumber\\
&=&  - \; \varepsilon
  \left(\frac{\partial E_p}{\partial v_{nm}^a}
 + 	\frac{\partial E_p}{\partial v_{nm}^b}{\sf i}
 + 	\frac{\partial E_p}{\partial v_{nm}^c}{\sf j}
 + 	\frac{\partial E_p}{\partial v_{nm}^d}{\sf k} \right), \ \ \\
\label{eqn2-2}
\Delta \gamma_n &:=& 
   \Delta \gamma_n^a
 + \Delta \gamma_n^b{\sf i} 
 + \Delta \gamma_n^c{\sf j} 
 + \Delta \gamma_n^d{\sf k} \nonumber\\
&=&  - \; \varepsilon 
  \left(\frac{\partial E_p}{\partial \gamma_n^a}
 + 	\frac{\partial E_p}{\partial \gamma_n^b}{\sf i} 
 + 	\frac{\partial E_p}{\partial \gamma_n^c}{\sf j} 
 + 	\frac{\partial E_p}{\partial \gamma_n^d}{\sf k}\right), \\
\label{eqn2-3}
\Delta w_{ml} &:=& 
   \Delta w_{ml}^a
 + \Delta w_{ml}^b{\sf i} 
 + \Delta w_{ml}^c{\sf j} 
 + \Delta w_{ml}^d{\sf k} \nonumber\\
&=&  - \; \varepsilon 
  \left(\frac{\partial E_p}{\partial w_{ml}^a}
 + 	\frac{\partial E_p}{\partial w_{ml}^b}{\sf i}
 + 	\frac{\partial E_p}{\partial w_{ml}^c}{\sf j} 
 + 	\frac{\partial E_p}{\partial w_{ml}^d}{\sf k} \right), \ \\
\label{eqn2-4}
\Delta \theta_m &:=& 
   \Delta \theta_m^a
 + \Delta \theta_m^b{\sf i} 
 + \Delta \theta_m^c{\sf j} 
 + \Delta \theta_m^d{\sf k} \nonumber\\
&=&  - \; \varepsilon 
  \left(\frac{\partial E_p}{\partial \theta_m^a}
 + 	\frac{\partial E_p}{\partial \theta_m^b}{\sf i} 
 + 	\frac{\partial E_p}{\partial \theta_m^c}{\sf j} 
 + 	\frac{\partial E_p}{\partial \theta_m^d}{\sf k} \right).
\end{eqnarray}
}
where $\Delta x$ denotes the amount of the correction of a parameter $x$. 
The above equations (\ref{eqn2-1}) - (\ref{eqn2-4}) can be expressed as 
%
\begin{align}
\label{eqn2-5}
    \Delta v_{nm}  = & \ \overline{H}_m \Delta \gamma_n,  \\[0.1cm]
\label{eqn2-6}
    \Delta \gamma_n  = & \ \varepsilon \Big( (1 - O^a_n)O^a_n\delta^a_n + (1 - O^b_n)O^b_n\delta^b_n \  {\sf i} \notag \\
                     & \quad + (1 - O^c_n)O^c_n\delta^c_n \ {\sf j} + (1 - O^d_n)O^d_n\delta^d_n \ {\sf k} \Big), 
\end{align}
%
%
\begin{align}
\label{eqn2-7}
    \Delta w_{ml} = & \ I_l {\Delta\overline{\theta}_m},  \\[0.1cm]
\label{eqn2-8}
    \Delta \theta_m  = & \ (1 - H^a_m)H^a_m \cdot Re       \left[\sum_n(\Delta{\overline{\gamma_n} v_{nm}})\right] \notag \\
                     & \quad + (1 - H^b_m)H^b_m \cdot Im^i \left[\sum_n(\Delta{\overline{\gamma_n} v_{nm}})\right]{\sf i}  \notag \\
                     & \quad + (1 - H^c_m)H^c_m \cdot Im^j \left[\sum_n(\Delta{\overline{\gamma_n} v_{nm}})\right]{\sf j}  \notag \\
                     & \quad + (1 - H^d_m)H^d_m \cdot Im^k \left[\sum_n(\Delta{\overline{\gamma_n} v_{nm}})\right]{\sf k}.
\end{align}
where 
$Re[x] := x_1, \; Im^i[x] := x_2,\; Im^j[x] := x_3$ and $Im^k[x] := x_4$ 
for a quaternion $x=x_1+x_2{\sf i} + x_3{\sf j} + x_4{\sf k} \in \mathbb{H}$.

%
For comparison, the amount of the correction for the usual quaternion 
neural network where $xw$ is calculated for a weight $w \in \mathbb{H}$ and an input signal $x \in \mathbb{H}$ to a quaternion neuron is shown below, which was 
derived in \cite{QuaternionBP}. 
\begin{eqnarray}
\label{eqn2-9}
\Delta v_{nm}   &=& \overline{H}_m \Delta \gamma_n,\\
\label{eqn2-10}
\Delta \gamma_n &=& \varepsilon \Big( \delta^a_n(1-O^a_n)O^a_n 
			       + \delta^b_n(1-O^b_n)O^b_n \ {\sf i} \nonumber\\
			   & & + \delta^c_n(1-O^c_n)O^c_n \ {\sf j} 
			       + \delta^d_n(1-O^d_n)O^d_n \ {\sf k} \Big), \\
%
\label{eqn2-11}
\Delta w_{ml}   &=& \overline{I}_l \Delta \theta_m,\\
\label{eqn2-12}
\Delta \theta_m &=& 
    (1-H^a_m)H^a_m \cdot Re \left[\sum_n(\Delta \gamma_n \overline{v}_{nm})\right]\nonumber\\   
& &+   (1-H^b_m)H^b_m \cdot Im^i\left[\sum_n(\Delta \gamma_n \overline{v}_{nm})\right]{\sf i} \nonumber\\
& &+ (1-H^c_m)H^c_m \cdot Im^j\left[\sum_n(\Delta \gamma_n \overline{v}_{nm})\right]{\sf j} \nonumber\\
& &+   (1-H^d_m)H^d_m \cdot Im^k\left[\sum_n(\Delta \gamma_n \overline{v}_{nm})\right]{\sf k}. \nonumber\\
\end{eqnarray}
The meaning of symbols in Eqs. (\ref{eqn2-9}) - (\ref{eqn2-12}) is the same as those in Eqs. (\ref{eqn2-5}) - (\ref{eqn2-8}).
The differences between the learning rule of the RQNN and the one of the usual quaternion 
neural network are the positions of $\delta_n^a, \delta_n^b, \delta_n^c, \delta_n^d$ in 
Eqs. (\ref{eqn2-6}) and (\ref{eqn2-10}), $\Delta \theta_m, I_l$ in Eqs. (\ref{eqn2-7}) 
and (\ref{eqn2-11}), and the quaternion conjugate for $\Delta \gamma_n, v_{nm}$ in Eqs. (\ref{eqn2-8}) and (\ref{eqn2-12}).


%
\section{Experiments}
In order to investigate the learning characteristics of the RQNN, we conducted experiments focusing on two rotation representations and evaluated their performance.
%

We call the multi-layer feedforward network with quaternion neurons, where $xw$ is calculated for a weight $w \in \mathbb{H}$ and an input signal $x \in \mathbb{H}$, \textit{QNN} for short.
%
\subsection{Rotation of Lines Task} \label{subsecA}
\begin{figure}[H]
    \raggedleft
    \includegraphics[width=92mm, height=55mm]{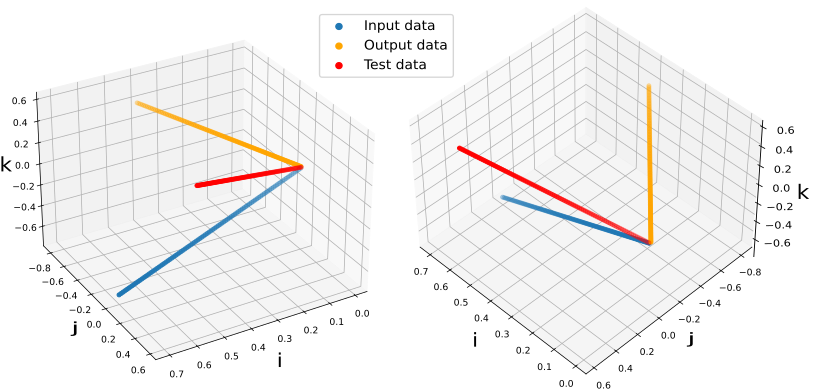}
    \caption{The blue represents the input learning data, 
    the yellow the output target data, and the red the input test data. 
    The figure on the right is a view of the figure on the left from a different angle.}
    \label{fig:rotation_test}
\end{figure}

We evaluated the RQNN's generalization performance for rotations through line rotation simulations.
In the three-dimensional space defined by the \textsf{i}, \textsf{j}, and \textsf{k} axes, we set a reference line along the diagonal direction $[1,1,1]/\sqrt{3}$ and sampled 99 points at 0.01 intervals from 0.01 to 0.99. This was used as the input test data.
Furthermore, we created a line along $[1,-2, 1]/\sqrt{6}$ that is perpendicular to the reference line, 
which served as the output target data, and another line along $[1,0,-1]/\sqrt{2}$ as the input learning data 
(Fig. \ref{fig:rotation_test}).
%

Both the RQNN and the QNN models were implemented as fully connected three-layer neural networks with a 4-8-4 architecture. 
The weights of the QNN and the RQNN were initialized from the same standard normal distribution to ensure a fair comparison. The sigmoid function was employed as the activation function for the hidden layer, while the identity function was used for the output layer. The learning rate was set to $5\times10^{-3}$, and training was continued until the training loss value decreased below 0.05.
%
To assess the learning characteristics of both models, we evaluated the differences in their outputs using the following four distinct evaluation metrics. The primary objective of this research is not to compare performance on test data, but rather to analyze the differences in internal learning characteristics resulting from different quaternion multiplication orders. Therefore, instead of conventional accuracy metrics, we employed these metrics to capture the differences in the internal representations of the models.
\vspace{0.1cm}
%
\begin{enumerate}
    \item \textbf{Angle Differences}: The angle between the tangent vectors at each point of the output curves of the QNN and the RQNN is calculated and converted from radians to degrees through multiplication by $180^\circ/\pi$, as defined by Eqs.(\ref{eq-theta}) and (\ref{eq-diff_angles}).
    \begin{align}
        \label{eq-theta}
        &\theta_i = \arccos \!\Bigl(
          \frac{\mathbf{t}^Q_i \cdot \mathbf{t}^{RQ}_i}
               {\|\mathbf{t}^Q_i\|\|\mathbf{t}^{RQ}_i\|}
        \Bigr),\\
        \label{eq-diff_angles}
        &\text{AngleDiff}
        = \frac{180^\circ}{\pi (N-1)} 
          \sum_{i=1}^{N-1} \theta_i.
    \end{align}
    Here, $\mathbf{Q}_i, \mathbf{RQ}_i \in \mathbb{R}^3$ denote the position vectors of the $i$-th output point of the QNN and the RQNN respectively, where $i \in \{1,\ldots,N\}$ and $N$ is the total number of points. The tangent vectors $\mathbf{t}^Q_i, \mathbf{t}^{RQ}_i \in \mathbb{R}^3$ are defined at the $i$-th point as $\mathbf{t}^Q_i = \mathbf{Q}_{i+1} - \mathbf{Q}_i$ for the QNN and $\mathbf{t}^{RQ}_i = \mathbf{RQ}_{i+1} - \mathbf{RQ}_i$ for the RQNN.
    \vspace{0.1cm}
    \item \textbf{Trajectory Differences}: The distance between the two curves is calculated using the Euclidean distance, as defined by Eq.(\ref{eq-Euclidean}), quantifying the overall difference in the curve shapes.
    \begin{align}
        \label{eq-Euclidean}
        &\text{TrajDiff} = \frac{1}{N}\sum_{i=1}^N \|\mathbf{Q}_i - \mathbf{RQ}_i\| \notag \\ 
             &\qquad + 0.1 \cdot \frac{1}{N-2}\sum_{i=1}^{N-2}|\kappa^Q_i - \kappa^{RQ}_i|. \\ 
        \setlength{\belowdisplayskip}{2.0ex}
        &\text{where}\: \kappa^m_i = \frac{\|\mathbf{v}^m_i \times \mathbf{a}^m_i\|}{\|\mathbf{v}^m_i\|^3}, \quad m \in \{Q, RQ\} \notag
    \end{align}
    \vspace{0.1cm}
    Here, $\kappa^m_i \in \mathbb{R}$ denotes the curvature at the $i$-th point for model $m \in \{Q, RQ\}$. The first derivative $\mathbf{v}^m_i \in \mathbb{R}^3$ is defined as $\mathbf{v}^m_i = \mathbf{Q}_{i+1} - \mathbf{Q}_i$ for $m = Q$ and $\mathbf{v}^m_i = \mathbf{RQ}_{i+1} - \mathbf{RQ}_i$ for $m = RQ$. The second derivative $\mathbf{a}^m_i \in \mathbb{R}^3$ is computed as $\mathbf{a}^m_i = \mathbf{v}^m_{i+1} - \mathbf{v}^m_i$.
    \vspace{0.1cm}
    \item \textbf{Phase Space Differences}: The differences in the first and second derivatives between the QNN and the RQNN are evaluated as defined by Eq.(\ref{eq-Phase_diff}). This metric combines the differences in first derivatives with a weighted term for second derivatives to quantify the dynamic characteristics of the curves.
    \begin{align}
        \label{eq-Phase_diff}
        \text{PhaseDiff} &= \frac{1}{N-1}\sum_{i=1}^{N-1} \|\mathbf{v}^Q_i - \mathbf{v}^{RQ}_i\| \notag \\ 
        & \qquad+ 0.1 \cdot \frac{1}{N-2}\sum_{i=1}^{N-2} \|\mathbf{a}^Q_i - \mathbf{a}^{RQ}_i\|
    \end{align}
    
    \item \textbf{Geometric Consistency}: The deviation of the angles between the tangent vectors from 90 degrees is calculated as defined by Eq.(\ref{eq-Geo_Consis}). The angle calculation in radians is converted to degrees through multiplication by $180^\circ/\pi$, measuring the geometric relationship between the test data and the outputs of both models.
    \begin{align}
       \label{eq-Geo_Consis}
       \text{GeoConsist} &= \frac{1}{2}\Bigg\{\sum_{m \in \{Q,RQ\}}\frac{1}{N-1}\sum_{i=1}^{N-1} \notag \\
       &\quad\left|\arccos\!\Bigl(\frac{\mathbf{v}^m_i \cdot \mathbf{t}_{test,i}}{\|\mathbf{v}^m_i\| \, \|\mathbf{t}_{test,i}\|}\Bigr) \cdot \frac{180^\circ}{\pi} - 90^\circ\right| \Bigg\}
    \end{align}
    Here, $\mathbf{t}_{test,i}$ denotes the tangent vector of the test data at the $i$-th point, calculated as $\mathbf{t}_{test,i} = \mathbf{T}_{i+1} - \mathbf{T}_i$, where $\mathbf{T}_i \in \mathbb{R}^3$ represents the position vector of the $i$-th point on the test data curve.
\end{enumerate}
\vspace{0.5cm}

\begin{figure}
    \centering
    \includegraphics[width=90mm, height=85mm]{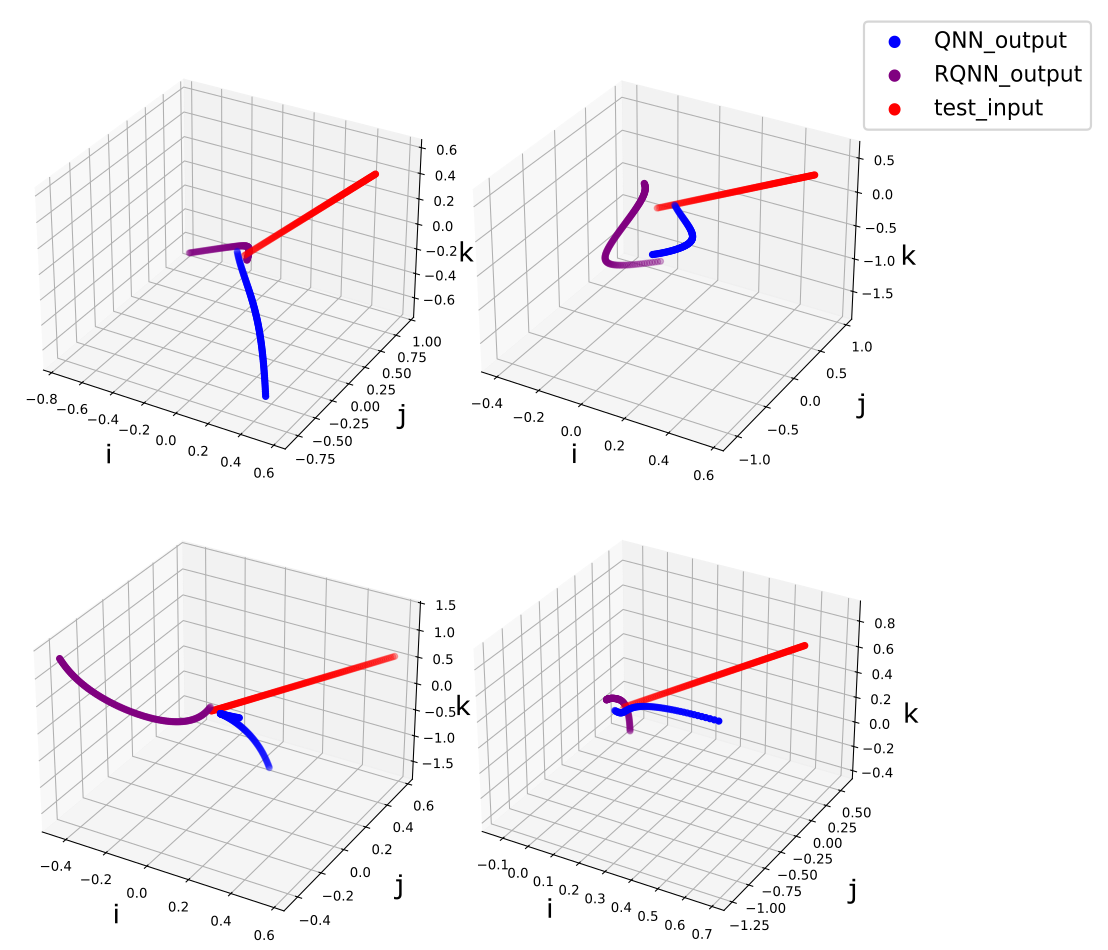}  
    \caption{Visualization of four experimental trials out of fifty trials. 
    The blue line shows the output of the QNN for the input test data (red line) after training.
    The purple line shows the output of the RQNN for the input test data (red line) after training.}
    \label{fig:rotation_plot_result}
\end{figure}

\begin{figure}
    \centering
    \includegraphics[width=90mm, height=70mm]{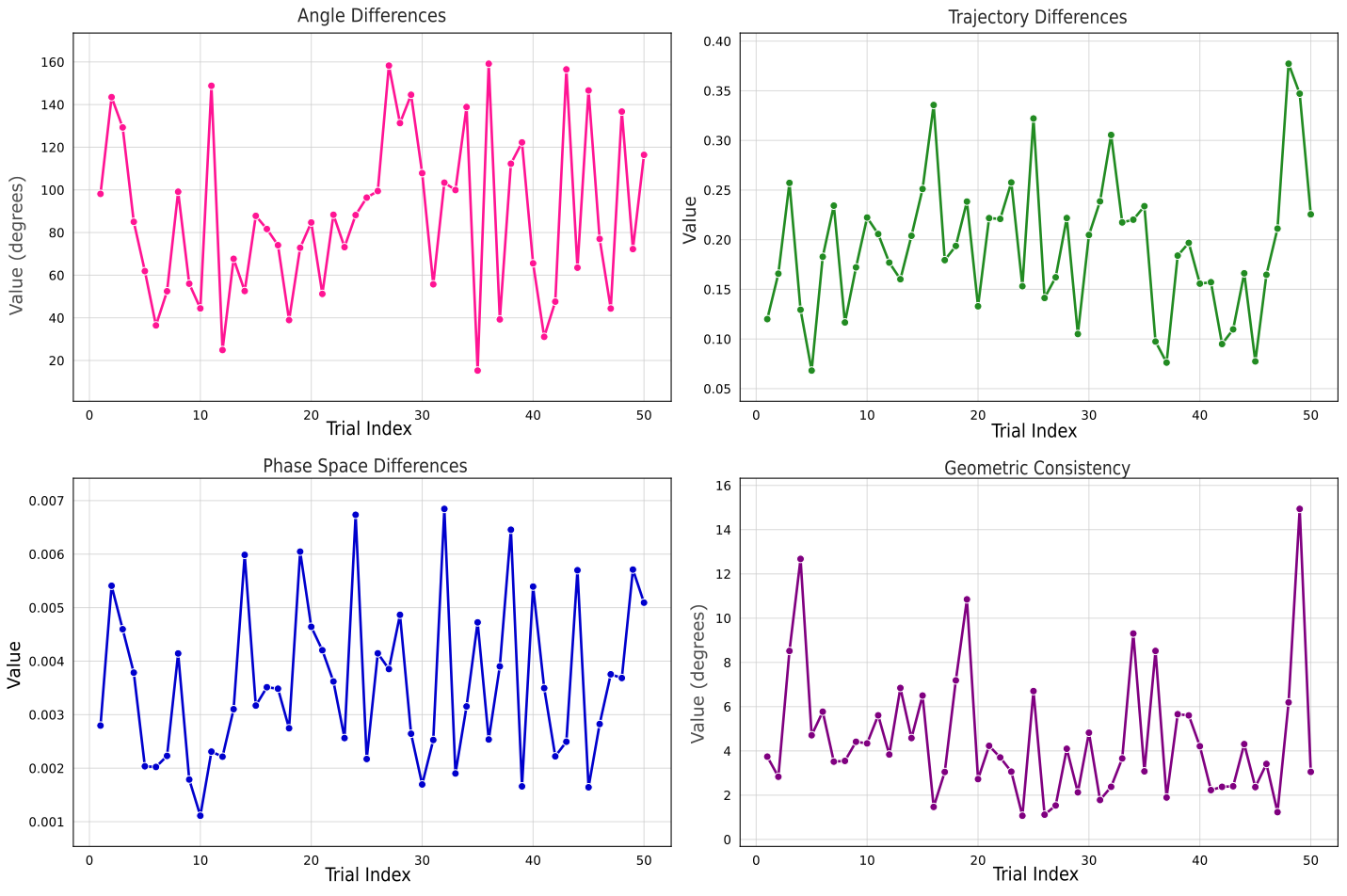}  
    \caption{Visualization of the evaluation metrics for the QNN and the RQNN outputs over 50 trials. The evaluation metrics used are as follows: (1) angle differences between tangent vectors, (2) trajectory differences (measured by Euclidean distance and curvature), (3) phase space differences in first and second derivatives, and (4) geometric consistency measured by deviation from 90 degrees orthogonality}

    \label{fig:4-evaluate}
\end{figure}

%
Fig. \ref{fig:rotation_plot_result} shows a representative example of the experimental results.
The blue and purple lines represent the outputs of the QNN and the RQNN respectively when the test data (shown in red) is provided as input after training.
As shown in Fig. \ref{fig:rotation_plot_result}, the QNN and the RQNN generate completely different output patterns despite being given the same test data as input.
Fig. \ref{fig:4-evaluate} shows the evolution of four evaluation metrics over 50 trials for both the QNN and the RQNN outputs, computed using Eqs.(\ref{eq-theta}) - (\ref{eq-Geo_Consis}).
Table \ref{tb-evaluation50} summarizes the statistical analysis results of these four evaluation metrics across the 50 trials.

As shown in Figs. \ref{fig:rotation_plot_result} and \ref{fig:4-evaluate} and Table \ref{tb-evaluation50}, the QNN and the RQNN exhibited clearly different learning behaviors, despite having identical initial weights and test data. Notably, the angle differences between the models' tangent vectors demonstrated a significant discrepancy, averaging 87.65 degrees.
%
%
%
%
\begin{table}
    \centering
    \begin{tabular}{lcccc}
       \hline
       Metric & Mean & Std & Min & Max \\
       \hline
       Angle Differences (degrees) & \textbf{87.65} & 38.73 & 15.27 & 159.16 \\
       Trajectory Differences & 0.19 & 0.07 & 0.07 & 0.37 \\
       Phase Space Differences & 0.0036 & 0.0015 & 0.0011 & 0.007 \\
       Geometric Consistency (degrees) & \textbf{4.56} & 2.89 & 1.070 & 14.93 \\
       \hline
   \end{tabular}
   \vspace{0.1cm}
   \caption{Statistical analysis of four evaluation metrics computed from Eqs.(\ref{eq-theta}) - (\ref{eq-Geo_Consis}) over 50 trials}
   \label{tb-evaluation50}
\end{table}

%


\subsection{Three-dimensional Object Rotation Task} 
Next, we investigated the learning characteristics of the RQNN for 3D point cloud object rotation using 3D point cloud datasets. In this experiment, we utilized the ModelNet10 dataset \cite{Wu2015}, which is a subset of ModelNet40. The ModelNet10 dataset consists of ten categories: chair, table, toilet, sofa, bed, desk, dresser, nightstand, bathtub, and monitor, and has been widely adopted as a standard benchmark for 3D object recognition.


%
The input data underwent random rotation, and as shown in Fig. 5, these rotated samples were used as the input learning data, while the original objects served as the output target data. This configuration enabled the model to learn the transformation from arbitrary rotated orientations to the original pose. The rotation matrices were generated using Rodrigues' rotation formula, with rotation angles constrained to a maximum of 90 degrees.

\begin{figure}[h]
    \centering
    \includegraphics[width=100mm, height=50mm]{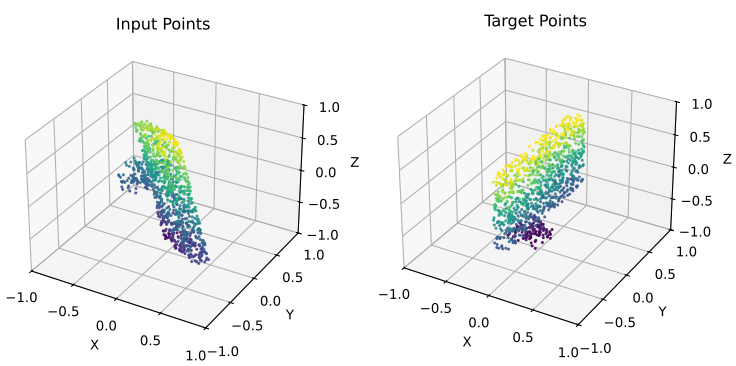}  
     \caption{Visualization of 3D point cloud data: Input learning points (left) and output target points (right). The rotated object serves as the input learning data, while the original, unrotated object serves as the output target data. This approach enables the model to learn a rotation that restores the object to its original orientation.}
    \label{fig:5-modelnet_feature}
\end{figure}

\begin{figure*}[h]
    \centering
    \includegraphics[width=195mm, height=65mm]{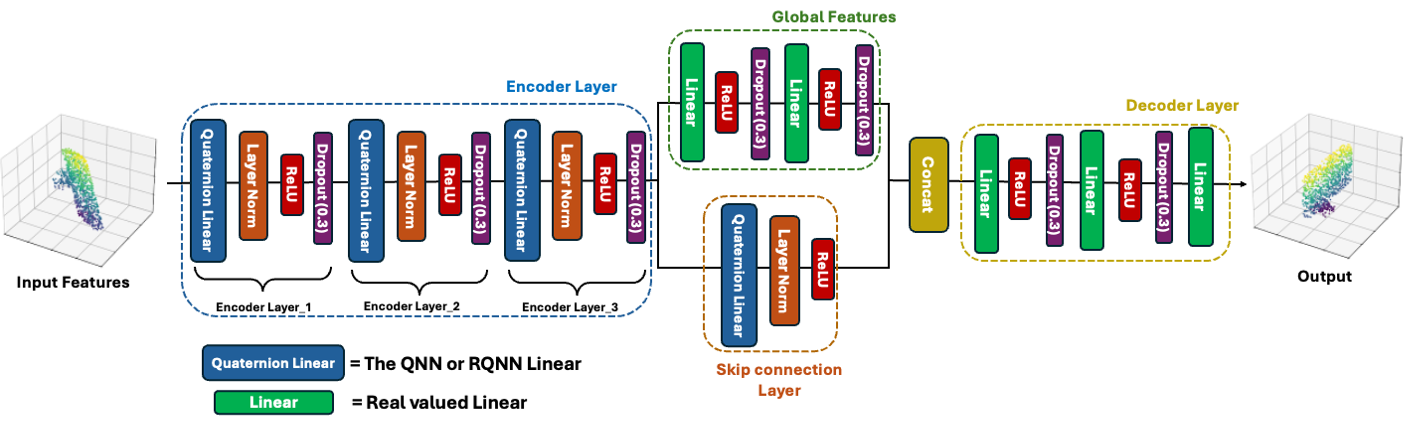}  
    \caption{Architecture of the proposed quaternion neural network model, consisting of a quaternion encoder for feature extraction, a global feature generator for higher-order representation learning, and a decoder with skip connections for 3D coordinate prediction.}
    \label{fig:6-model_architecture}
\end{figure*}

%
Using this data, we developed models incorporating the RQNN and the QNN as architectural components and investigated their learning behavior by analyzing weights, activations, and gradients after training.
%
%
We constructed a model architecture incorporating the QNN and the RQNN modules within the encoder structure (Fig. \ref{fig:6-model_architecture}). The model consists of three main components: an encoder for feature extraction, a global feature generator, and a decoder. The input three-dimensional data is converted to pure quaternions by setting the real component to zero and assigning the data to the imaginary components, resulting in 4-dimensional quaternion features. The encoder comprises three quaternion fully-connected layers that perform quaternion multiplication operations, sequentially transforming these features through 16, 32, and 16 dimensions. Each quaternion linear layer is followed by normalization, ReLU activation, and dropout.

The global feature generator achieves higher-order representations by flattening the encoder output ($16$ dimensions $\times$ number of input points) and processing it through two fully-connected layers ($16 \times 1024 \rightarrow 1024 \rightarrow 512$ dimensions). To enhance model expressiveness and preserve local feature information, we implemented skip connections from the encoder's final layer ($16$ dimensions).

The decoder concatenates global features ($512$ dimensions) with skip connection features ($16$ dimensions) to generate a $528$-dimensional feature vector, which is then processed through three real-valued fully-connected layers ($528 \rightarrow 256 \rightarrow 128 \rightarrow 3$ dimensions) to predict the final 3D coordinates.

%

%
The loss function was constructed as a weighted sum of MSE loss, Chamfer distance, and geometric loss, computed according to Eqs.(\ref{eq-mse_loss}) - (\ref{eq-total_loss}). Each loss term serves a specific purpose in the training process: the MSE loss ensures point-wise accuracy, the Chamfer distance handles global point cloud alignment by measuring bidirectional point-to-point distances, and the geometric loss preserves local structural relationships by considering neighboring point configurations. This combination allows the model to maintain both global shape consistency and local geometric details during the rotation transformation.

\begin{align}
\label{eq-mse_loss}
\text{MSE} &= \frac{1}{N}\sum_{i=1}^N \|x_i - y_i\|^2,  \\
\label{eq-Chamfer_loss}
\text{Chamfer} &= \frac{1}{B}\sum_{b=1}^B\left(\frac{1}{N}\sum_{i=1}^N\min_j\|x_i - y_j\|_2 \right. \notag \\ 
           &\qquad \left. + \frac{1}{N}\sum_{j=1}^N\min_i\|x_i - y_j\|_2\right),  \\
\label{eq-Structure_loss}
\text{Structure} &= \frac{1}{NK}\sum_{i=1}^N\sum_{k=1}^K\|(x_i - x_{i,k}) - (y_i - y_{i,k})\|^2,  \\[1em]
\label{eq-total_loss}
\text{Loss} &= \text{MSE} + \alpha \text{Chamfer} + \beta \text{Structure} .
\end{align}
where $x_i$ and $y_i$ represent predicted and target point coordinates respectively, $N$ denotes the number of points, $j \in \{1,\ldots,N\}$ is the index for target points, $B$ is the batch size, $K$ represents the number of nearest neighbors, $\alpha$ and $\beta$ are the weights for Chamfer and geometric losses respectively, and $\|\cdot\|_2$ denotes the L2 norm (Euclidean distance). The subscripts $i,k$ in $x_{i,k}$ and $y_{i,k}$ indicate the $k$-th nearest neighbor of the $i$-th point.

\begin{figure}
    \centering
    \includegraphics[width=90mm, height=85mm]{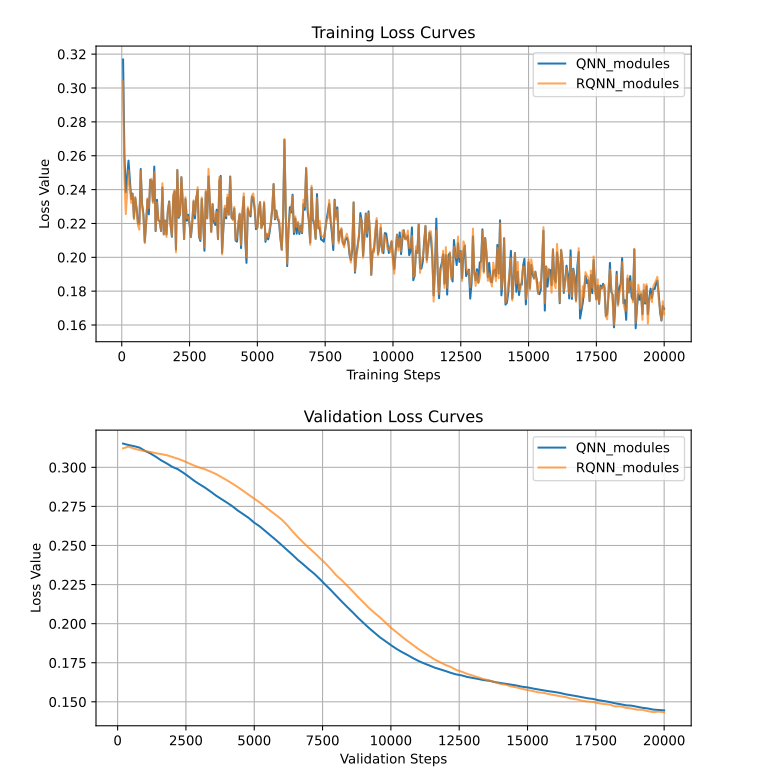}  
    \caption{Learning curves for models incorporating QNN and RQNN modules. Both architectures demonstrate similar convergence characteristics, exhibiting comparable behavior throughout the training process.}
    \label{fig:7-total_loss_curve}
\end{figure}

%
%
The model comprises 17.5M parameters and was trained using the Adam optimizer with a batch size of 16 and a learning rate of $1\times10^{-6}$. The loss weights $\alpha$ and $\beta$ were set to 0.1 and 0.2 respectively, with the number of nearest neighbors $K$ set to 20. Twenty percent of the training data was reserved for validation, and the model was trained for 100 epochs. The learning curves for both models are shown in Fig. \ref{fig:7-total_loss_curve}, and the final loss values computed using Eq.~(\ref{eq-total_loss}) for the test data are presented in Table~\ref{tb-test_total_loss}.

%
%
\begin{table}[h]
   \centering
   \begin{tabular}{lcc}
      \hline
      Model & Params & Loss  \\
      \hline
      QNN-modules  & 17.5M & 0.1499  \\
      RQNN-modules & 17.5M & 0.1501 \\
      \hline
  \end{tabular}
  \vspace{0.1cm}
  \caption{Comparison of loss values between QNN and RQNN-integrated models evaluated on test data}
  \label{tb-test_total_loss}
\end{table}

\begin{figure}[h]
    \centering
    \includegraphics[width=90mm, height=85mm]{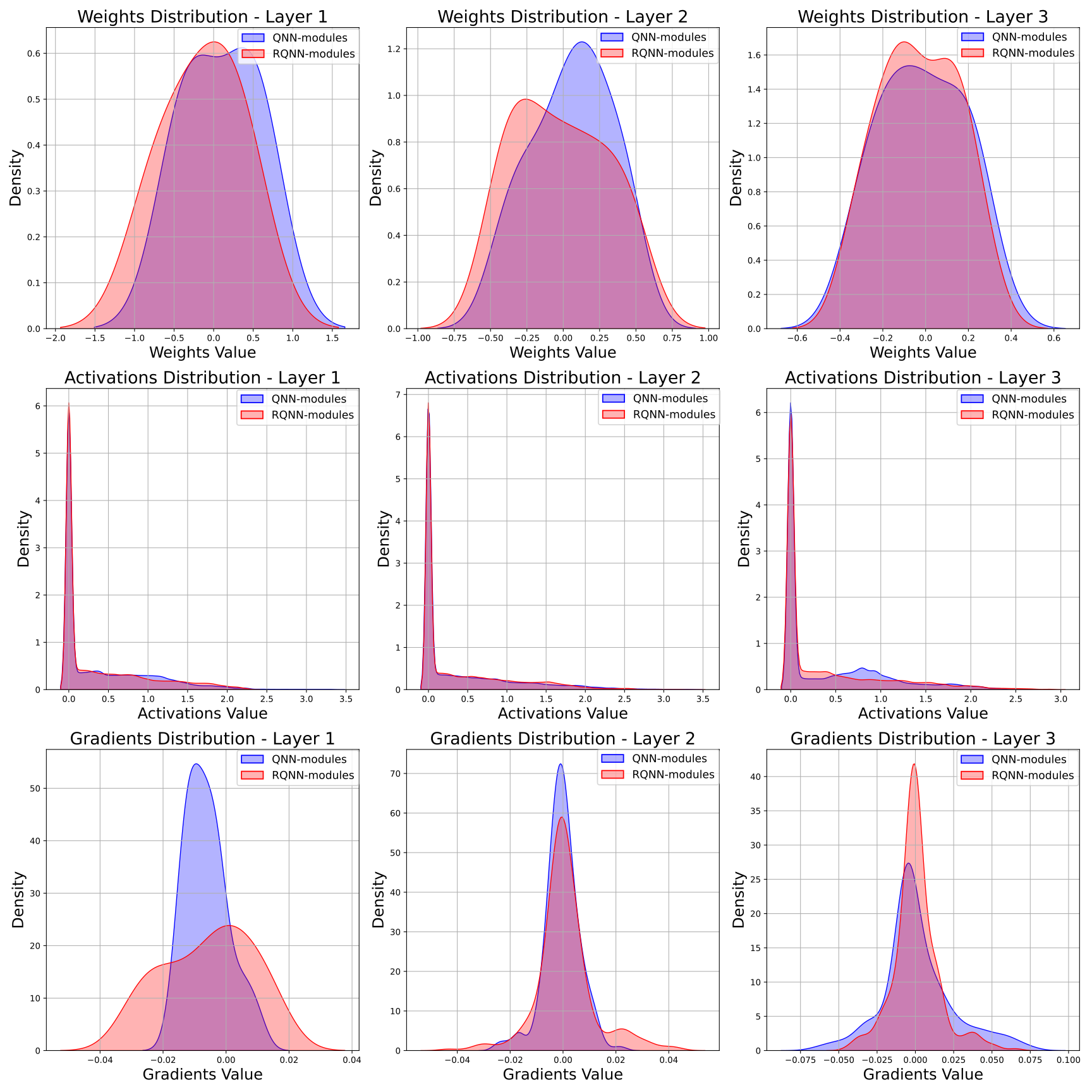}  
    \caption{Comparative analysis of models with QNN and RQNN modules. Distributions of weights (top), activations (middle), and gradients (bottom) are visualized across three encoder layers, demonstrating distinctive patterns in parameter space exploration and learning dynamics between the two architectural approaches.}
    \label{fig:8-garaidents_plot}
\end{figure}

%
Using the trained models, we extracted the encoder sections composed of the QNN and the RQNN architectures and analyzed the learning behavior of the three quaternion fully-connected layers for both models on the test data. The distributional characteristics of weights, activations, and gradients were visualized through Kernel Density Estimation (KDE) plots in Fig. \ref{fig:8-garaidents_plot}, and their statistical measures are summarized in Table \ref{tb-pretrain_model_statics}.

%

%
\begin{table*}[h]
\centering
\begin{tabular}{|cc| c c| c c| c c|}
\hline
Layer & Model & Weight\_Avg & Weight\_Std & Activation\_Avg & Activation\_Std & Gradient\_Avg & Gradient\_Std \\
\hline
1 & QNN & 0.094638 & 0.415134 & 0.411645 & 0.578605 & -0.006139 & 0.006907 \\
1 & RQNN & -0.137196 & 0.443018 & 0.398682 & 0.565451 & -0.005593 & 0.013652 \\
\hline
2 & QNN & 0.056367 & 0.270802 & 0.394833 & 0.625221 & -0.000385 & 0.006376 \\
2 & RQNN & -0.018313 & 0.314268 & 0.401765 & 0.599941 & 0.000887 & 0.010724 \\
\hline
3 & QNN & -0.017184 & 0.196021 & 0.415738 & 0.568344 & 0.001543 & 0.022889 \\
3 & RQNN & -0.030565 & 0.181208 & 0.395756 & 0.600589 & 0.000568 & 0.015029 \\
\hline
\end{tabular}

\vspace{0.1cm}
\caption{Statistical distributions of weights, activations, and gradients across layers in the QNN and the RQNN}
\label{tb-pretrain_model_statics}
\end{table*}

%
As illustrated in Fig. \ref{fig:7-total_loss_curve}, models incorporating the QNN and the RQNN modules achieved convergence with nearly identical learning trajectories. While the test data results in Table \ref{tb-test_total_loss} indicate marginally better performance for the QNN model, the difference in performance metrics between the two approaches remains negligible.
In contrast, analysis of Fig. \ref{fig:8-garaidents_plot} and Table \ref{tb-pretrain_model_statics} reveals that while both models achieved similar performance metrics, they exhibited distinct learning characteristics in feature acquisition during training, as evidenced by their different gradient patterns and weight distributions.

%
\section{Discussion}
%
%
%
In the rotation of lines task, the learning behavior of the RQNN exhibited distinct differences compared to conventional models.
As shown in Fig. \ref{fig:rotation_plot_result}, the outputs for identical test data demonstrated significant variations. 
Analysis of Angle Differences from Table \ref{tb-evaluation50} and Fig. \ref{fig:4-evaluate} revealed that despite some variability, the average across 50 trials was 87.65 degrees, indicating both models successfully learned 90-degree rotations. The Geometric Consistency averaging 4.56 further corroborates the high-precision achievement of 90-degree rotation learning.
The Trajectory Differences averaged 0.19, indicating a consistent disparity between the two outputs. While the Geometric Consistency showed a spike around the 48th trial, it maintained stability with an average of 4.56 degrees and a standard deviation of 2.89, demonstrating the high-precision achievement of 90-degree rotation learning. The Phase Space Differences exhibited remarkably small values, with an average of 0.0036 and a standard deviation of 0.0015, indicating that both models share similar operational characteristics in their dynamic behavior.
These findings indicate that while the RQNN shows differences from the QNN in terms of output orientation and trajectory, it maintains the fundamental operational characteristics.

In the 3D object rotation task using ModelNet10, both models incorporating the QNN and the RQNN modules successfully achieved learning objectives, exhibiting remarkably similar learning curves.
As shown in Table \ref{tb-test_total_loss}, the loss values on test data were nearly identical, indicating comparable performance between the two models. 
However, when comparing the quaternion layers extracted from the trained models, significant differences in weight distributions and gradient behavior were observed between the two architectures, as illustrated in Fig. \ref{fig:8-garaidents_plot} and Table \ref{tb-pretrain_model_statics}.
Regarding weight distributions, the RQNN weights in Layer 1 and Layer 2 showed mean values of $-0.137$ and $-0.018$, with standard deviations of $0.44$ and $0.31$ respectively, demonstrating a broader distribution compared to the QNN weights. This spread suggests that the RQNN potentially explores a wider range of features during the early stages of learning. In Layer 3, both models' weight distributions converged to similar shapes as training progressed. The observed bias towards negative mean values in the weights can be attributed to the differences in quaternion multiplication order.
The activation distributions exhibited highly similar patterns across all layers for both the QNN and the RQNN, suggesting that the order of quaternion multiplication has minimal impact on activation values. However, significant differences were observed in gradient distributions between the two models. the QNN demonstrated a tendency to efficiently direct learning towards specific directions from the early stages. In contrast, the RQNN's gradient distributions initially showed a wider range before concentrating as training progressed. This wider distribution characteristic of the RQNN potentially offers advantages in terms of learning stability and gradient vanishing prevention, while gradually refining its feature representation in deeper layers.

%
\section{Conclusion} 
In this paper, we proposed a new neuron structure for multi-layer feedforward quaternion neural networks called the Reverse Quaternion Neural Network (RQNN), which leverages the non-commutative property of quaternion multiplication, and investigated its learning characteristics. Through comparative analysis with conventional models in lines rotation tasks, we found that the RQNN exhibited distinct generalization capabilities. Furthermore, our experiments with 3D object rotation tasks using ModelNet10 demonstrated that while the QNN and the RQNN architectures achieved similar performance metrics, they developed fundamentally different learning representations and strategies. 

%

Future research directions should focus on deeper investigation of RQNN's learning characteristics and representation properties. While our current study demonstrated unique generalization properties through fundamental experiments, comprehensive exploration across diverse application domains remains necessary. Although we examined RQNN's learning behavior through line rotation and ModelNet10 experiments, validation across various datasets and more complex object structures is essential. These investigations will help identify specific domains where RQNN's unique characteristics, particularly the differences in generalization capabilities from existing models and its broader feature exploration in early learning stages, can be effectively leveraged for enhanced performance. 
%
The authors also feel that the work presented in this paper is
probably only scratching the surface of what is possible
by focusing on the inherent properties of hypercomplex algebras. 

%
\subsubsection*{Acknowledgments.}
The authors would like to thank the anonymous reviewers for valuable comments. This paper is part of the master's thesis that Shogo Yamauchi, one of the authors, 
completed in 2024 at the Graduate School of Artificial Intelligence and Science, 
Rikkyo University.

%

\begin{thebibliography}{8}

\bibitem{hirose2013}
Hirose, A.: Complex-Valued Neural Networks: Advances and Applications. IEEE Press/Wiley (2013)

\bibitem{lee2022}
Lee, C., Hasegawa, H., Gao, S.: Complex-valued neural networks: A comprehensive survey. IEEE/CAA Journal of Automata Sinica \textbf{9}(8), 1--21 (2022)

\bibitem{mandic2009}
Goh, V.S.L., Mandic, D.P.: Complex Valued Nonlinear Adaptive Filters: Noncircularity, Widely Linear and Neural Models. Wiley, New York (2009)

\bibitem{nit2009}
Nitta, T.: Complex-Valued Neural Networks: Utilizing High-Dimensional Parameters. Information Science Reference, Pennsylvania, USA (2009)

\bibitem{parcollet2020}
Parcollet, T., Morchid, M., Linarès, G.: A survey of quaternion neural networks. Artificial Intelligence Review \textbf{53}, 
2957--2982 (2020). Springer, \doi{10.1007/s10462-019-09752-1}

\bibitem{complexNN}
Nitta, T.: An Extension of the Back-Propagation Algorithm to Complex Numbers. Neural Networks \textbf{10}, 1391--1415 (1997)

\bibitem{QuaternionBP}
Nitta, T.: A quaternary version of the back-propagation algorithm. In: Proceedings of ICNN'95 - International Conference on Neural Networks, vol. 5, pp. 2753-2756 vol.5 (1995). \doi{10.1109/ICNN.1995.488166}

\bibitem{nit1991}
Nitta, T., Furuya, T.: A complex back-propagation learning. Transactions of Information Processing Society of Japan \textbf{32}(10), 1319--1329 (1991). In Japanese

\bibitem{nit1993}
Nitta, T.: A complex numbered version of the back-propagation algorithm. In: Proc. INNS World Congress on Neural Networks (WCNN1993), vol. 3, pp. 576--579 (1993), Portland, July 11-15

\bibitem{zhu2019quaternion}
Zhu, X., Xu, Y., Xu, H., Chen, C.: Quaternion Convolutional Neural Networks. In: Proceedings of the European Conference on Computer Vision (ECCV), September (2018)

\bibitem{lecun1998gradient}
LeCun, Y., Bottou, L., Bengio, Y., Haffner, P.: Gradient-based learning applied to document recognition. Proceedings of the IEEE \textbf{86}(11), 2278--2324 (1998)

\bibitem{parcollet2018quaternion}
Parcollet, T., Zhang, Y., Morchid, M., Trabelsi, C., Linarès, G., De Mori, R., Bengio, Y.: Quaternion Convolutional Neural Networks for End-to-End Automatic Speech Recognition. In: Proc. Interspeech 2018, pp. 22--26 (2018). \doi{10.21437/Interspeech.2018-1898}

\bibitem{Quaternion}
Hamilton, W.R.: ON QUATERNIONS, OR ON A NEW SYSTEM OF IMAGINARIES IN ALGEBRA By William Rowan Hamilton, \url{https://www.maths.tcd.ie/pub/HistMath/People/Hamilton/OnQuat/OnQuat.pdf}, last accessed 2023/10/25

\bibitem{matsui2004quaternion}
Matsui, N., Isokawa, T., Kusamichi, H., Peper, F., Nishimura, H.: Quaternion neural network with geometrical operators. Journal of Intelligent \& Fuzzy Systems \textbf{15}(3-4), 149--164 (2004)

\bibitem{Hopfield-type_QNN}
Yoshida, M., Kuroe, Y., Mori, T.: Models of Hopfield-type quaternion neural networks and their energy functions. International journal of neural systems \textbf{15}, 129--135 (2005)


\bibitem{rumel}
Rumelhart, D.E., Hinton, G.E., Williams, R.J.: Learning internal representations by error propagation. In: Parallel Distributed Processing: Explorations in the Microstructures of Cognition, vol. 1, pp. 318--362. MIT Press, Cambridge, MA (1986)

\bibitem{Wu2015}
Z. Wu, S. Song, A. Khosla, F. Yu, L. Zhang, X. Tang, and J. Xiao :  3D ShapeNets: A deep representation for volumetric shapes
in \textit{2015 IEEE Conference on Computer Vision and Pattern Recognition (CVPR)}, Jun.2015, pp.1912--1920.
\end{thebibliography}


%


\end{document}